# A LEARNING-BASED VISUAL SALIENCY FUSION MODEL FOR HIGH DYNAMIC RANGE VIDEO (LBVS-HDR)

*Amin Banitalebi-Dehkordi[1], Yuanyuan Dong[1], Mahsa T. Pourazad[1,2], and Panos Nasiopoulos[1]*

[1]ECE Department and ICICS at the University of British Columbia, Vancouver, BC, Canada
[2]ICICS at the University of British Columbia & TELUS Communications Incorporation, Canada

## ABSTRACT

Saliency prediction for Standard Dynamic Range (SDR) videos has been well explored in the last decade. However, limited studies are available on High Dynamic Range (HDR) Visual Attention Models (VAMs). Considering that the characteristic of HDR content in terms of dynamic range and color gamut is quite different than those of SDR content, it is essential to identify the importance of different saliency attributes of HDR videos for designing a VAM and understand how to combine these features. To this end we propose a learning-based visual saliency fusion method for HDR content (LVBS-HDR) to combine various visual saliency features. In our approach various conspicuity maps are extracted from HDR data, and then for fusing conspicuity maps, a Random Forests algorithm is used to train a model based on the collected data from an eye-tracking experiment. Performance evaluations demonstrate the superiority of the proposed fusion method against other existing fusion methods.

## 1. INTRODUCTION

When watching natural scenes, a large amount of visual data is delivered to Human Visual System (HVS). To efficiently process this information, the HVS prioritizes the scene regions based on their importance, and performs in-depth processing on the regions according to their associated priority [1]. Visual Attention Models (VAMs) evaluate the likelihood of each region of an image or a video to attract the attention of the HVS. Designing accurate VAMs has been of particular interest for computer vision scientists as VAMs not only mimic the layered structure of the HVS, but also provide means to dedicate computational resources for video/image processing tasks efficiently. In addition VAMs are used for designing quality metrics as they allow quantifying the effect of distortions based on the visual importance of the pixels of an image or video frame.

Over the last decade, saliency prediction for Standard Dynamic Range (SDR) video has been well explored. However, despite the recent advances in High Dynamic Range (HDR) video technologies, limited amount of work on HDR VAM exists [2, 3]. Considering that the SDR VAMs are designed for SDR video and do not take into account the wide luminance range and rich color gamut associated with HDR video content, they fail to accurately measure the saliency for the HDR video [2, 3].

VAMs are mostly designed based on Feature Integration Theory [4]. Various saliency attributes (usually called conspicuity or feature maps) from image or video content are extracted and combined to achieve an overall saliency prediction map. To combine various feature maps to a single saliency map, it is a common practice in the literature to use linear averaging [5]. Different weights may be assigned to different features. In a study by Itti et al. [5] to combine various feature maps, a Global Non-Linear Normalization followed by Summation (GNLNS) is utilized. GNLNS normalizes the feature maps and emphasizes on local peaks in saliency. Unfortunately it is not clearly known that how the HVS fuses various visual saliency features to assess an overall prediction for a scene. Motion, brightness contrast, color, and orientation have been identified as important visual saliency attributes in literature [2, 5-8]. Considering that the characteristic of HDR content in terms of dynamic range and color gamut is quite different than SDR content, it is important to investigate the importance of these saliency attributes for designing a HDR VAM.

In this paper, our objective is to model how the HVS fuses various visual saliency features of HDR video content and identify the importance of each feature. In our implementation, we extract motion, color, intensity, and orientation saliency features as suggested in Dong's bottom-up HDR saliency prediction approach [2]. Once the feature maps are extracted, we use a Random Forests (RF) algorithm [9] to train a model using the results of an eye-tracking experiment over a large database of HDR videos (watched on a HDR prototype display). This model efficiently combines different HDR saliency features, and provides robustness and flexibility to add new features, or reduce the features and keep the most important features through feature importance analysis. The effectiveness of the proposed Learning Based Visual Saliency (LBVS-HDR) fusion model is demonstrated through objective metrics and visual comparison with eye-tracking data.

This work was partly supported by Natural Sciences and Engineering Research Council of Canada (NSERC) under Grant STPGP 447339-13 and Institute for Computing Information and Cognitive Systems (ICICS) at UBC.

The rest of this paper is organized as follows: Section 2 elaborates on our methodology, Section 3 explains the specifications of the HDR video database and the eye-tracking experiments procedure, Section 4 includes the evaluation results and discussions, while Section 5 concludes the paper.

## 2. METHODOLOGY

This section first elaborates on various saliency features used in our study, and then describes our proposed Learning Based Visual Saliency fusion model.

### 2.1. HDR saliency attributes

The HDR saliency features used in our approach are motion, color, brightness intensity, and texture orientations, as suggested by the state of the art VAMs for SDR [5] and HDR [2]. For taking into account the color and luminance perception under wider HDR luminance range, before the feature channels are extracted, the HDR content is processed using the proposed HDR HVS modeling module by Dong [2]. This HDR HVS modeling module consists of three different parts (see Fig. 1):

- Color Appearance Model (CAM): accounts for colors being perceived differently by the HVS under different lighting conditions [10-12]. The HDR HVS modeling module uses the CAM proposed by Kim et al. [11] as it covers a wide range of luminance (up to 16860 cd/m$^2$) [11]. Using this CAM, two color opponent signals (red-green (R-G) and yellow-blue (B-Y)) are generated, which then form the color saliency features (see [2, 11] for more details).
- Amplitude Nonlinearity: models the relationship between the physical luminance of a scene with the perceived luma by HVS (unlike the SDR video, the luminance is not modeled by a linear curve for HDR video). The luminance component of the HDR signal is mapped to a so-called luma values using Just Noticeable Difference (JND) mapping of [13].
- Contrast Sensitivity Function (CSF): accounts for the variations of the contrast sensitivity at different spatial frequencies. The HDR HVS modeling module uses the CSF model of Daly [14].

Once the HDR content is processed by Dong's HDR HVS modeling module [2], different parallel feature channels (namely motion, color, brightness intensity, and texture orientations) are extracted. As suggested in [3], for extracting motion feature map, an optical flow based approach is used (residual flow detection method of [15,16]) to ensure high performance in bright image areas and minimize the effect of brightness flickering between video frames. For each of the color, intensity and orientation features as proposed in [2, 5], a Gaussian pyramid is built in different scales and these pyramids are later combined across scales to form a feature map. Once feature maps are generated, they need to be fused to create a Spatio-temporal Map. In our study we propose to use a Random

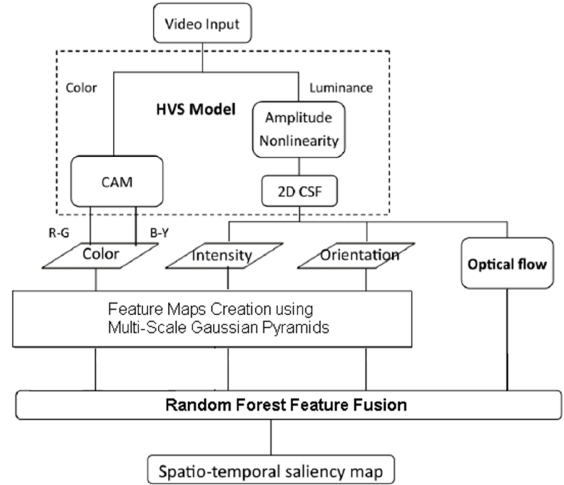

**Fig. 1.** Flowchart of Random Forests Fusion approach

Forests-based fusion approach to imitate HVS Spatial-temporal information fusion (see Fig. 1). The following subsection elaborates on our feature fusion approach.

### 2.2. Our feature fusion approach

The existing visual attention models, which are based on the Feature Integration Theory, extract several saliency features and combine them to one saliency map. As previously mentioned, most of the existing methods represent the overall saliency map as the average of generated features. However, it is not known exactly how this kind of fusion is performed by the HVS. It is likely that different saliency features to have different impact on the overall saliency. Therefore different weights should be assigned to them and these weights may vary spatially over each frame.

In this paper, we propose to train a model based on Random Forests algorithm for the fusing feature maps. By definition, RF is a classification and regression technique, which combines bagging and random feature selection to construct a collection of Decision Trees (DTs) with controlled variance [9]. Although the DTs do not perform well on unseen test data individually, the collective contribution of all DTs makes RF generalize well to unseen data. This is also the case for predicting visually important areas within a scene based on temporal and spatial saliency maps; while individual saliency features are not quite successful in predicting the visual importance of a scene, integration of these features provides a much more accurate prediction. Random Forests Regression is of particular interest in our study, as it only needs very little parameter tuning, while its performance is robust for our purpose. Moreover, the RF algorithm evaluates the importance of each feature in the overall prediction.

We train a model of Random Forests using the training part of our HDR video database and evaluate its performance using the validation video set. Details regarding the model creation procedure are provided in Section 3.

**Table 1.** Specifications of the HDR video database used for eye tracking experiments

| Sequence | Frame Rate | Resolution | Number of Frames | Source | Usage |
|---|---|---|---|---|---|
| Balloon | 30 fps | 1920x1080 | 200 | Technicolor [19] | Training |
| Bistro02 | 25 fps | 1920x1080 | 300 | Froehlich et al. [20] | Training |
| Bistro03 | 30 fps | 1920x1080 | 170 | Froehlich et al. [20] | Training |
| Carousel08 | 30 fps | 1920x1080 | 439 | Froehlich et al. [20] | Training |
| Fishing | 30 fps | 1920x1080 | 371 | Froehlich et al. [20] | Training |
| MainMall | 30 fps | 2048x1080 | 241 | DML-HDR [16] | Training |
| Bistro01 | 30 fps | 1920x1080 | 151 | Froehlich et al. [20] | Validation |
| Carousel01 | 30 fps | 1920x1080 | 339 | Froehlich et al. [20] | Validation |
| Market | 50 fps | 1920x1080 | 400 | Technicolor [19] | Validation |
| Playground | 30 fps | 2048x1080 | 222 | DML-HDR [16] | Validation |

## 3. EXPERIMENT SETUP

In order to train and test the model for fusing the temporal and spatial saliency features, we prepare a HDR video database and perform eye-tracking experiments using this database. The following subsections provide details regarding the HDR videos, eye-tracking experiments, and our RF fusion model estimation.

### 3.1. HDR videos

To the best of the authors' knowledge, to this date, there is no publicly available eye-tracking database of HDR videos. To validate the performance of our proposed Learning Based Visual Saliency Fusion model for HDR, we prepare an HDR video database and conduct eye tracking tests using this database. Ten HDR videos are selected from the HDR video database at the University of British Columbia [17-18], Technicolor [19], and Froehlich et al. [20]. The selection of the test material is done in a way that the database contains night, daylight, indoor, and outdoor scenes with different amounts of motion, texture, colorfulness, and brightness.

Next, the test HDR video database is divided to training and validation sets. Six for sequences are chosen for training and four for validation of the Random Forests model. Table 1 provides the specifications of the utilized HDR video database.

### 3.2. Eye tracking experiments

Test material was shown to the viewers using a Dolby HDR prototype display. This display system includes a LCD screen in front, which displays the color components and a projector at the back, which projects the luminance component of the HDR signal. The projector light converges on the LCD lens to form the HDR signal. Details regarding the display system are available in [21]. The resolution of the TV is 768×1024, the peak brightness of the overall system is 2700 cd/m$^2$, and the color Gamut is BT. 709.

A free viewing eye tracking experiment was performed using the HDR vides and the HDR display prototype. Eye gaze data was collected using SMI I View X RED device at the sampling frequency of 250 Hz and accuracy of 0.4±0.03$^o$. 18 subjects participated in our tests. For each participant, the distance and height was adjusted to ensure that the device is fully calibrated. All participants were screened for vision and color perception acuity.

The eye tracker automatically records three types of eye behavior: fixations, saccades and blinks. Fixations and information associated with each fixation are used to generate fixation density maps (FDMs). The FDMs represent subjects' region of interest (RoI) and serve as ground truth for assessing the performance of visual attention models. To generate FDMs for the video clips, spatial distribution of human fixations for every frame is computed per subject. Then, the fixations from all the subjects are combined together and filtered by a Gaussian kernel (with a radius equal to one degree of visual angle). More details on our Eye tracking experiment are provided in [22]. Our HDR eye-tracking database is publicly available at [23].

### 3.3. RF-based fusion

In our study to fuse the temporal and spatial maps, we train a model of Random Forests. The RF model is estimated using the extracted feature maps of the HDR training data set (see Table 1) as input and the corresponding eye Fixation Density Maps as output. For a fast implementation, we use only 10% of the training videos (equal number of frames selected from each video, we ensured to select representative frames from each scene). We choose 100 trees, boot strap with sample ratio (with replacement) of 1/3, and a minimum number of 10 leaves per tree. Note that these parameters are chosen for demonstration purposes (our complementary experiments showed that using higher percentage of training data provides better performance, but at the price of higher computational complexity). Once the RF fusing model is trained, the saliency map of unseen HDR video sequences – validation video set - is predicted based on their temporal and spatial saliency maps.

## 4. RESULTS AND DISCUSSIONS

To compute importance of each feature in the model-training phase the out-of-bag error calculation is used [9]. As observed from Table 2, motion feature achieves the highest importance. We also evaluate the performance of each feature map individually in the saliency prediction.

**Table 2.** Relative feature importance

| Feature | Importance |
|---|---|
| Motion | 1 |
| Color | 0.96 |
| Orientation | 0.83 |
| Intensity | 0.50 |

**Table 3.** Individual performance of different features using the validation video set

| Feature (alphabetical order) | AUC | sAUC | EMD | SIM | PCC | KLD | NSS |
|---|---|---|---|---|---|---|---|
| Motion | 0.63 | 0.62 | 0.09 | 0.36 | 0.23 | 0.31 | 1.27 |
| Color | 0.61 | 0.59 | 0.19 | 0.34 | 0.19 | 0.28 | 0.97 |
| Orientation | 0.60 | 0.57 | 0.07 | 0.33 | 0.14 | 0.36 | 0.67 |
| Intensity | 0.56 | 0.56 | 0.11 | 0.30 | 0.08 | 0.30 | 0.40 |

**Table 4.** Performance evaluation of different feature fusion methods

| Fusion Method | AUC | sAUC | EMD | SIM | PCC | KLD | NSS |
|---|---|---|---|---|---|---|---|
| Average | 0.63 | 0.60 | 0.12 | 0.35 | 0.20 | 0.32 | 0.78 |
| Multiplication | 0.58 | 0.57 | 0.08 | 0.29 | 0.12 | 1.42 | 0.49 |
| Maximum | 0.60 | 0.59 | 0.19 | 0.33 | 0.16 | 0.39 | 0.64 |
| Sum plus product | 0.63 | 0.60 | 0.12 | 0.35 | 0.20 | 0.33 | 0.80 |
| GNLNS (Itti [5]) | 0.64 | 0.62 | 0.11 | 0.35 | 0.21 | 0.33 | 0.85 |
| Least Mean Squares Weighted Average | 0.62 | 0.60 | 0.13 | 0.34 | 0.18 | 0.32 | 0.71 |
| Weighting according to STD of each map | 0.63 | 0.60 | 0.15 | 0.35 | 0.20 | **0.30** | 0.77 |
| **Random Forest** | **0.68** | **0.67** | **0.07** | **0.38** | 0.21 | **0.30** | **0.99** |

Table 3 contains the result of training and validation different RF models when only one feature map is used.

To evaluate the performance of the proposed LBVS-HDR method on the validation video set, we use several saliency evaluation metrics so that our results are not biased towards a particular metric. Specifically, we use the Area Under the ROC Curve (AUC) [24], shuffled AUC (sAUC) [24], Kullback–Leibler Divergence (KLD) [25], Earth Mover's Distance (EMD) [26], Natural Scan-path Saliency (NSS) [24], Pearson Correlation Ratio (PCC), and Judd et al. saliency similarity measure (SIM) [27]. Note that in each case, the metric values are calculated for each frame of the videos in the validation video set and then averaged over the frames. For all the metrics except for KLD and EMD, higher values represent better performance. Here we compare the performance of different saliency fusion schemes for saliency prediction using the proposed feature maps by [3]. We include the state-of-the-art fusion methods in our comparisons. Table 4 illustrates the performance of different feature fusion methods over the validation video set. As it is observed the RF fusion achieves the highest performance using different metrics. Table 5 demonstrates the results of various feature fusion schemes as well as the ground truth fixation maps for one of the validation sequences.

Our study shows that motion and color are highly salient features for the observers. In addition, our proposed LBVS-HDR fusion model is capable of efficiently combining various saliency feature maps to generate an overall HDR saliency map.

## 5. CONCLUSION

In this paper, we proposed a learning-based visual saliency fusion (LBVS-HDR) method for HDR videos. Several Saliency features adapted to the high luminance range and rich color gamut of HDR signals are extracted and efficiently combined using a Random Forests-based fusion model. Performance evaluations confirmed the superiority of our proposed fusion method against the existing ones. Also we found that motion and color are the most important attributes of the salient regions of HDR content.

**Table 5.** Comparison of different feature fusion methods

| Fusion Method | Carousel01 |
|---|---|
| Video Frame | 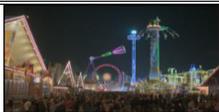 |
| Average | 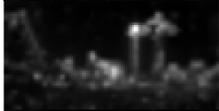 |
| Multiplication | 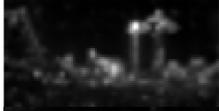 |
| Maximum | 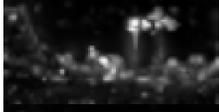 |
| Sum plus product | 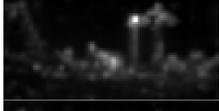 |
| GNLNS (Itti [5]) | 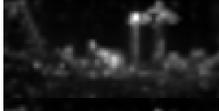 |
| Least Mean Squares Weighted Average | 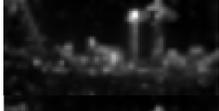 |
| Weighting according to STD of each map | 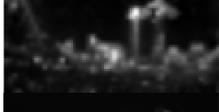 |
| **Random Forest** | 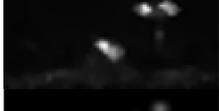 |
| Ground Truth | 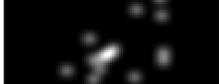 |